# Integrated design and fabrication of pneumatic soft robot actuators in a single casting step

A. Silva,[1] D. Fonseca,[1] D. Neto[1], M. Babcinschi[1], P. Neto[1]*

[1]University of Coimbra, CEMMPRE, ARISE, Department of Mechanical Engineering, Coimbra, Portugal. *Corresponding author. Email: pedro.neto@dem.uc.pt

**Abstract**

Bio-inspired soft robots have already shown the ability to handle uncertainty and adapt to unstructured environments. However, their availability is partially restricted by time-consuming, costly and highly supervised design-fabrication processes, often based on resource intensive iterative workflows. Here, we propose an integrated approach targeting the design and fabrication of pneumatic soft actuators in a single casting step. Molds and sacrificial water-soluble hollow cores are printed using fused filament fabrication (FFF). A heated water circuit accelerates the dissolution of the core's material and guarantees its complete removal from the actuator walls, while the actuator's mechanical operability is defined through finite element analysis (FEA). This enables the fabrication of actuators with non-uniform cross sections under minimal supervision, thereby reducing the number of iterations necessary during the design and fabrication processes. Three actuators capable of bending and linear motion were designed, fabricated, integrated and demonstrated as three different bio-inspired soft robots, an earthworm-inspired robot, a four-legged robot, and a robotic gripper. We demonstrate the availability, versatility and effectiveness of the proposed methods, contributing to accelerating the design and fabrication of soft robots. This study represents a step toward increasing the accessibility of soft robots to people at a lower cost.

**MAIN TEXT**

1. **Introduction**

    Soft robots provide a freedom of movement and flexibility similar to animals' soft bodies [1–5], making them an attractive choice for innovative robot designs and applications in unstructured environments that were previously unattainable using rigid-bodied robots [6]. Nevertheless, despite soft robot's appealing properties and recent multidisciplinary advances [7–9], further research is needed to improve their availability, feasibility of design and fabrication [10–12]. Research on soft robotics has yielded multiple morphologies and actuation principles to create bio-inspired robotic solutions [13–16]. Fluidic actuation is the most commonly used principle, characterized by its relatively fast

and efficient actuation [17, 18]. Multiple methods have been studied to assist in the design of fluidic soft actuators, the FEA of hyperelastic materials is one of the most promising tools to predict performance and optimize design and operability parameters [19–22].

The fabrication of fluidic-powered soft actuators with geometrically complex non-uniform cross section chambers and channels poses several scientific and technological challenges. The traditional multi-step casting of two-part silicones and their variants remains the most widely used fabrication process due to its widespread availability [23–25]. However, it is a highly supervised process, requiring skilled labor to conduct multiple processing and assembly steps, which may lead to noticeable defects when joining various silicone parts cured at different time periods. On the other hand, single-step cast molding is limited to producing actuators with uniform cross section chambers, as this is the only way to extract the core after curing the elastomer [26]. These limitations in chamber design significantly restrict the achievable morphological complexity of the actuator and, consequently, the robot's natural looking motion. The same applies to bubble casting, which has proved capable of high-volume production, but it is limited to continuous geometries with uniform cross sections [27]. The lost-wax casting process was demonstrated to be a promising technique. However, it is prone to dimensional inaccuracies due to thermal retraction, and the wax needs to be melted, which can leave unwanted residue on the silicone surfaces. Soluble materials have been introduced to fabricate sacrificial mold cores with uniform or non-uniform cross sections [28–30]. Materials such as polyvinyl alcohol (PVA) and acrylonitrile butadiene styrene (ABS) can be dissolved in water and acetone, respectively [31–32]. Using a solvent, the core can be removed to create the actuator's chambers. While this process is readily available, allowing for the fabrication of cores using FFF, the quality of printing is highly dependent on the printing parameters and environmental conditions. Additionally, the complete removal of the soluble material can be challenging and time-consuming to achieve in the presence of 3D intricate chamber geometries.

In recent years, we have witnessed major advances in the 3D printing fabrication of soft actuators, aimed at achieving the ultimate goal of 3D printing fully functional robots [33–38]. Fabrication technologies such as Stereolithography (SLA), Selective Laser Sintering (SLS), Digital Projection Lithography (DLP), Continuous Liquid Interface Production (CLIP) and Two-Photon Polymerization (2PP) are characterized by high feature resolution and smooth surface finish. Inkjet-based technologies such as PolyJet and MultiJet are capable of 3D printing fully functional robots, fabricating soft actuators with integrated fluidic circuits, printing them in a single run and using multiple soft/rigid materials [34], some of them at the microscale size [39, 40]. While these processes accelerate the fabrication of soft robots requiring minimal supervision, their availability is limited due to the high cost of some equipment (two to three orders of magnitude higher than desktop FFF printers), the cost of materials, and the limited selection of materials for use [35, 41]. Moreover, some materials can be toxic and may necessitate postprocessing to remove support structures. Nevertheless, recent studies have reported successful 3D printing of soft robots without the need for expensive equipment and materials, using FFF and SLS-based technologies [42–45]. Direct Ink Writing (DIW) has also enabled the successful fabrication of soft actuators and sensors [46–50]. Adding sacrificial "fugitive inks" to DIW allows the fabrication of robots with complex networks of microfluidic channels [10, 51], and print objects that react to magnetic fields and thermal stimulus [52, 53]. A similar technique, liquid-solid co-printing, has also been used for printing multi-material fluidic circuits for both hydraulic and combustion-powered soft robots [33, 54, 55]. Multimaterial

Multinozzle 3D printing (MM3D) enables the monolithic printing of pneumatic actuated soft robots, allowing the simultaneous fabrication of both actuators and sensors [56, 57].

Soft materials have evolved in tandem with advances in fabrication methods. Stimulus-responsive materials lead the way towards 4D printing [58, 59]. Self-healing hydrogels with adjustable properties hold the promise of being highly effective and flexible in a wide range of soft robotics applications [60]. An ultraviolet-curable elastomer used for DLP printing has proven capable of 1,100% elongation and application in the development of thermoelectric actuators [61, 62]. Photocurable inks for DIW have also enabled the fabrication of highly stretchable self-healing components [63]. FFF-based technology makes it possible to print soft structures using commercially available thermoplastic elastomers [42, 64, 65], hydrogels [66] and magnetic composites [67].

Here, we propose fabricating soft actuators using widely available and affordable processes, combining single-step cast molding with the FFF printing of sacrificial water-soluble cores. The actuator's mechanical operability is defined through FEA using a nonlinear hyperplastic material model. Although the use of sacrificial mold cores is common in the fabrication of soft actuators, this process is highly dependent on the chamber geometry and requires specific conditions such as the solvent temperature and flow, among other factors. We propose a heated water circuit to speed up the dissolution of the hollow core's material, ensuring complete removal from the actuator's walls, even for intricate chamber geometries, Fig. 1A-D. The process was validated and demonstrated through the integrated design-fabrication of three pneu-net inspired actuators [68] featuring bending and linear motion capabilities upon pressurization. These actuators were incorporated into three bio-inspired robots. An earthworm-inspired robot, built with a single linear actuator, moves as a result of the state-dependent friction between the convex shape of its legs and the ground, Fig. 1E. Four bending actuators, each one acting as a statically stable leg, enable the locomotion of the four-legged robot across different terrains, Fig. 1F. A robotic gripper, composed of three bending actuators, can grasp objects of varying weight, size, geometry and surface texture, Fig. 1G.

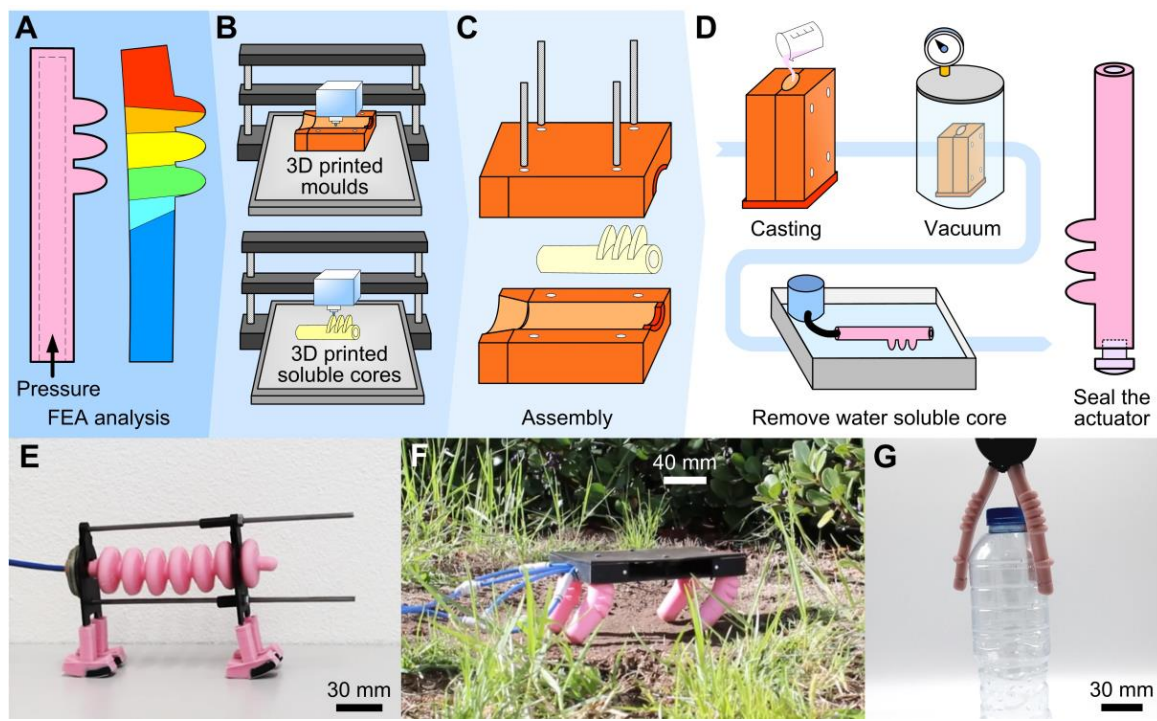

**Fig. 1**. Design-fabrication of soft robot actuators in a single casting step. (A) FEA-based numerical simulation to optimize the actuator's geometry and actuation parameters. (B) Using FFF technology, the molds are printed in rigid material and the sacrificial cores in a water-soluble material. (C) Molds and core are assembled following a standard molding process. (D) The elastomer in the liquid state is poured into the molds, followed by a period in the vacuum chamber to remove air bubbles. Once the elastomer is cured and extracted from the molds the heated water circuit is used to dissolve the sacrificial water-soluble cores, and finally, the actuator is sealed. (E) Earthworm-inspired robot composed of a single linear actuator. (F) Four-legged robot moving on a flat garden surface. (G) A soft pneumatic gripper grasping a plastic water bottle. All robots are pneumatically actuated.

## 2. Materials and Methods

### 2.1 Actuator fabrication

Single-step cast molding proved effective in fabricating soft fluidic actuators. However, due to the need to extract the core after elastomer curing, its applicability is limited to actuators with chambers that have a uniform cross section, frequently a simple tubular geometry. Here, this issue is addressed by utilizing sacrificial cores made of a water-soluble material, PVA, which can be fabricated using standard FFF technology. The casting process ends with the water-soluble core inside the actuator, Fig. 2A-B. The problem lies in the fact that achieving the complete removal of the water-soluble core can be a time-consuming and challenging process. It is time-consuming because the dissolution of the material occurs at a slow pace due to the generally small contact area between the material and the solvent. It is challenging because it is difficult to fully remove the soluble material from the actuator walls, especially in intricate geometries, Fig. S1.

We found that the water-soluble core can be completely removed in a short amount of time when submerged in a heated water bath and subjected to a flow of water passing through the core, Fig. 2C. The cores were specifically designed and fabricated with a through hole to enable the circulation of heated water. This design choice increased the contact area of the water with the core, thereby accelerating the dissolution of the core's material. The continuous flow of filtered water was maintained by pumping water through the core hole using a water pump. A resistance heats the water to aid the dissolution. The heated water provides energy to the system, facilitating the process of breaking the intermolecular forces in PVA for a more effective dissolution. The PVA manufacturer recommends temperatures higher than 50ºC. Our system maintains the water temperature at approximately 65ºC to strike a balance between achieving a higher temperature without compromising the operability of elements in contact with the heated water, specifically the pump which is rated to operate at temperatures lower than 70ºC. This method implies that the actuators must be designed with two open channels for the water to flow through the hollow core, with one of them being sealed afterward. The sealing can be accomplished in three different ways: (i) using an end cap made of non-soft material, (ii) using an end cap made of soft material glued to the actuator, or (iii) employing a casting process. In the latter case, it becomes a two-step casting process. The versatility of the process was demonstrated by successfully fabricating different soft actuators, namely, a linear actuator and two bending actuators, Fig. 2D and Movie S1, Supplementary Materials. The actuator's internal walls are free of soluble material, without leaks and the need for manual post-treatments, Fig. 2E and Fig. S1, Supplementary Materials. On the other hand, two primary defects were identified: geometrical inaccuracies manifested through non-uniform wall thickness and surface defects on the chamber walls, Fig. 2E. These defects arise from the high sensitivity of PVA water-soluble material to exposure to moisture and humidity, adding complexity to the FFF-based printing process of the cores. During the printing, problems such as popping, bubbling, poor bonding between layers, and inconsistent extrusion were identified, potentially resulting in rough surface finish and deformations on the printed cores. The core's rough surface finishing, along with intricate chamber geometries, makes core removal more challenging and time-consuming. Our experiments demonstrated that 20 minutes is sufficient to completely remove the water-soluble material. The above defects do not compromise the functionality of the actuators.

Overall, our hypothesis was confirmed, demonstrating that an effective process for removing the printed water-soluble cores makes it possible to speed up the fabrication of soft actuators using a single casting step.

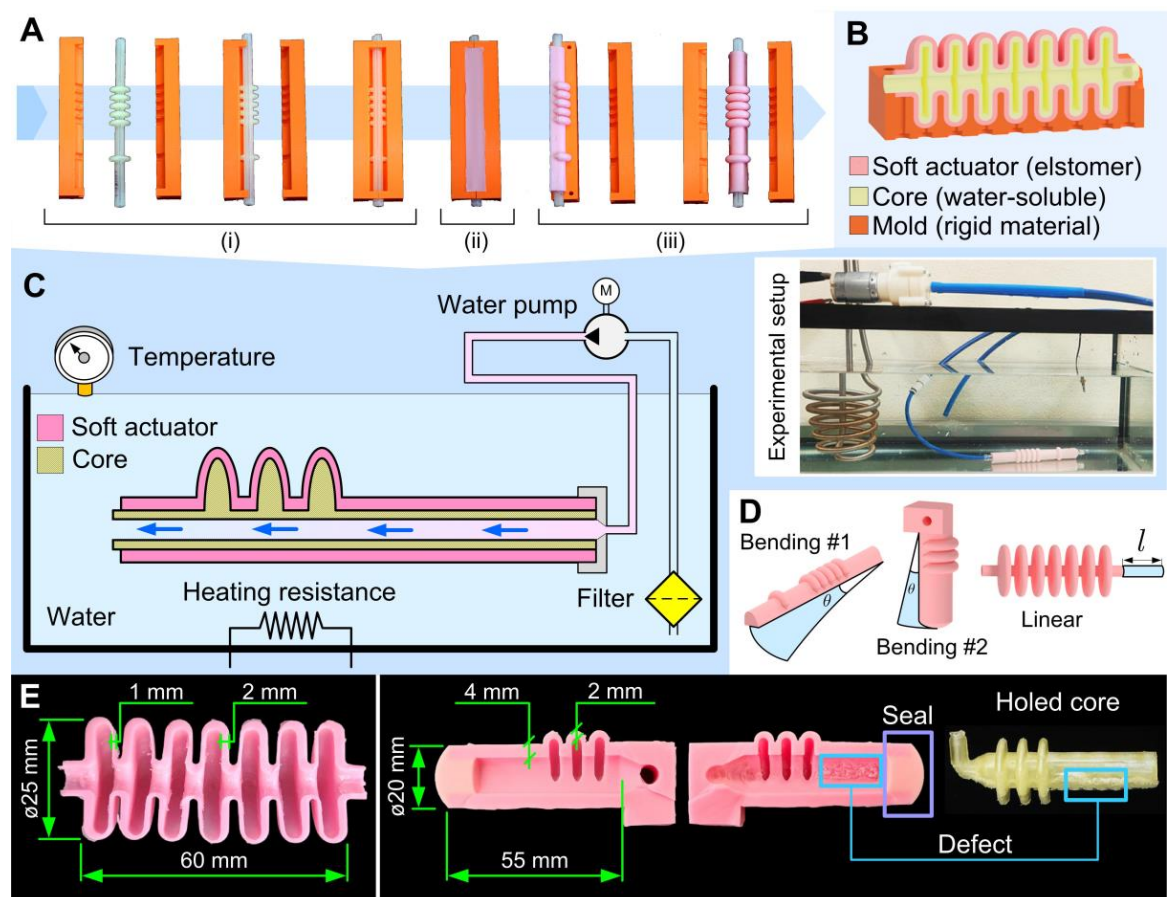

**Fig. 2**. Actuator fabrication and heated water circuit to remove sacrificial water-soluble cores. (A) Fabrication steps: (i) assembling the water-soluble sacrificial core in the molds, (ii) single casting process, and (iii) extracting the cured elastomer with the core inside from the molds. (B) Section view of an actuator showing the mold, the actuator and the hollow core. (C) Schematic of the heated water circuit in which the water flows through the hollow core, facilitating its removal. The actuator is placed in a water tank heated by a resistance and monitored by a temperature sensor, while a pump forces the circulation of filtered water through the hollow core. (D) Three exemplary actuator models featuring bending and linear motion capabilities. (E) Defects observed in the fabricated actuators include geometrical inaccuracies, such as non-uniform wall thickness of the linear actuator, and surface defects on the chamber walls of the bending #2 actuator. The dimensions refer to the actuators used in the robot prototypes, but they can be fabricated in different dimensions as required.

## 2.2 Materials and fabrication procedure

The actuators were made of platinum-catalyzed silicone (SmoothSil 940, Smooth-On, USA) featuring a Shore A hardness of 40, a 100% modulus of 1.38 MPa, and a tensile strength of 4.14 MPa. Water-soluble cores were 3D printed using biodegradable and non-toxic PVA filaments (SMARTFIL PVA, Smart Materials 3D, Spain). The rigid structural elements, including molds and robot components, were 3D printed using polylactic acid (PLA) (Prusa, Czechia). The heated water tank was filled with tap water.

Each actuator was fabricated following the sub-processes outlined in Fig. 1B-D and Movie S1, Supplementary Materials. The elastomer in liquid state was poured into the molds, degassed in a vacuum chamber (VC2509AG, VacuumChambers, Poland) for a period of ~2 min, cured at atmospheric conditions (20 to 25 ºC, 1 atm) for around 24 hours, extracted from the molds, and then the heated water circuit dissolved the sacrificial water-soluble cores. Finally, the actuator was sealed using an end cap made of soft material glued to the actuator. The molds, water-soluble cores and robot components were all 3D printed on a commercially available FFF 3-axis single-nozzle machine (Prusa i3 MK3S+, Prusa, Czechia). The heated water circuit consists of a 180×60×75 cm glass tank, a water pump to circulate the water (R385, Gaotu, China) and a 2000 W resistance for heating the water (2000W, Eurojava, Portugal). The system was controlled by a microcontroller (Nano, Arduino, Italy) receiving temperature feedback from a temperature sensor (DS18B20, Analog Devices, USA). The printer's G-code was generated from a slicer software (PrusaSlicer 2.5.0, Prusa, Czechia), using the computer-aided design (CAD) model of the components as input. Actuators, molds, cores and robots were modeled using CAD software (Inventor 2021, Autodesk, USA). Detailed visual information about molds, cores, and fabricated actuators is provided in Fig. S1, Supplementary Materials.

## 2.3 Finite element analysis

The mechanical behavior of the actuators is dictated by different design parameters, such as geometry, dimensions and material. On the other hand, their operability is mainly defined by the relationship between the pressure in the internal chamber and the deformed configuration of the actuator. We defined the actuators' geometry and main dimensions according to our robots' requirements, while using FEA simulations (Inventor Nastran, Autodesk, USA) to establish the relationship between the input pressures and the actuator's elongation/bendability. The mechanical behavior of the silicone elastomer was defined by the hyperelastic Neo-Hookean model. The strain energy density function for an incompressible material is defined as $W = A_{10}\left(D_1 - 3\right)$, where $A_{10}$ is the material constant parameter and $D_1$ is the first invariant of the right Cauchy-Green deformation tensor. The geometry of the actuators was modeled using quadratic tetrahedral finite elements. The finite element size was set at half of the nominal thinnest geometric feature of the actuator (1 mm). The input pressure was modeled as a boundary load applied to the internal surfaces of the actuators. This load was incrementally applied in 300 steps to account for the changing pressure area, which increases during loading. To compare the numerical and experimental data, experimental tests were conducted at room temperature (20 to 25 ºC, 1 atm) and recorded using a camera (EOS 1300D 18-55IS, Canon, Japan). The actuators ground truth displacements and bending angles were determined by analyzing the static image frames recorded by the camera.

The material parameter defining the constitutive model was calibrated using an iterative procedure that compares numerical predictions and physical measurements. The agreement between numerical and experimental data made it possible to validate the model, which was subsequently used in the analysis of the elongation/bending of each actuator dictated by the input pressure. The parameter $A_{10}$ was defined to be 0.24 MPa.

Considering different values of input pressure, the elongation of the linear actuator predicted numerically is in good agreement with the experimental data, Fig. 3A. The difference between the numerical and experimental maximum displacement is lower than 13%, Fig. 3D. Here, due to the actuators' relatively thin wall, the radial expansion is noticeable, restraining the linear displacement. The actuators' bending #1 and bending #2 present similar numerical and experimental radial deflections due to the bending effect, Fig. 3B and Fig. 3C, respectively. For both actuators, the difference between the numerical and experimental measurement of the maximum radial displacement was 2% for bending #1 and 8% for bending #2, Fig. 3D and Movie S2, Supplementary Materials. As expected, there is a relationship between the thickness of the wall and the resulting bending angle for the same input pressure. Bending #1 is lighter and longer, presenting a thinner wall, reaching a bending angle of ~60° when subjected to a pressure of 60 kPa, while bending #2 reaches an angle of ~15° at the same pressure in experimental conditions, Fig. 3E. However, the design of bending #2 provides more strength and stability, a key factor for the design of statically stable soft legs that hold their position under static conditions. In general, we found that FEA is an efficient tool to aid the design of soft actuators, eliminating lengthy and costly trial-and-error design-fabrication processes.

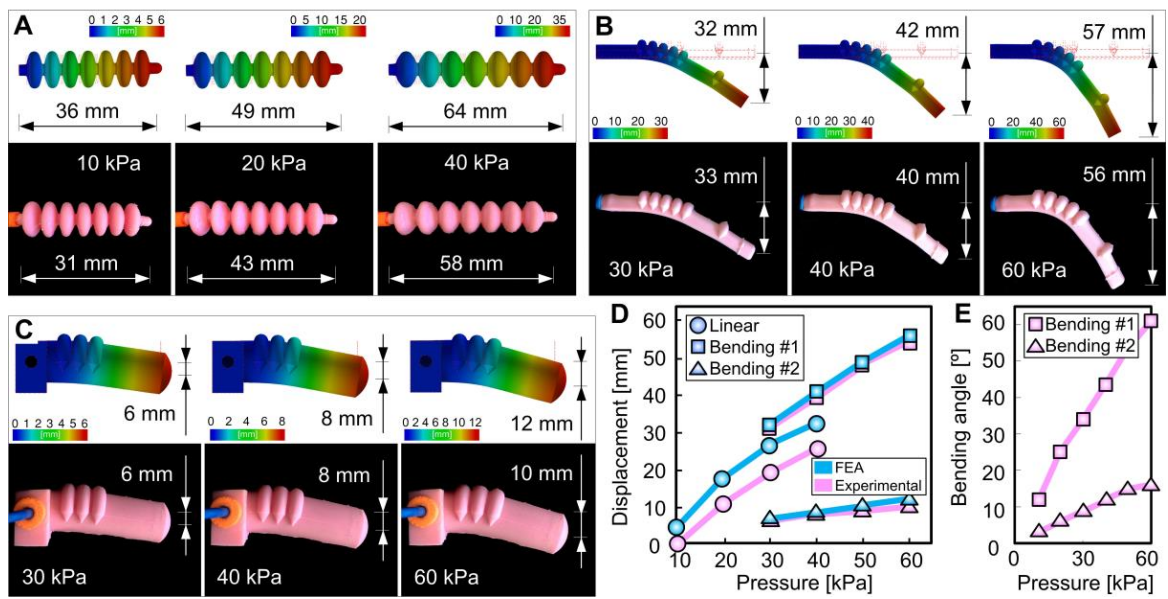

**Fig. 3**. Numerical predictions versus experimental measurements. (A) Elongation of the linear actuator for different values of input pressures. (B) Radial deflection of the actuator bending #1 (vertical displacement) for different values of input pressure. (C) Radial deflection of the actuator bending #2 (vertical displacement) for different values of input pressure. (D) Maximum displacement reached by the actuators for different pressure values. The blue curve represents the numerical results, whereas the pink curve represents the experimental ones. (E) The experimental bending angles of the actuators' bending #1 and bending #2 for different pressure values.

## 2.4 Monitoring and control elements

We utilized an off-board portable air compressor equipped with a pressure regulator to generate and control the airflow to the actuators (TE-AC 270/50/10, Einhell, Germany). The valves, sensors and microcontroller were powered by a programmable DC power supply (72-13360, TENMA, China). Two 6V solenoid 2/2 valves (CY05820D, cydfx, China) were used to direct the airflow to each actuator, Fig. S2, Supplementary Materials. The actuator's pressure was monitored by an absolute pressure sensor (MPX4250AP, NXP Semiconductors, Netherlands) operating at a sampling rate of 50 Hz. The microcontroller (Nano, Arduino, Italy) executed the on-off controllers, receiving data from the sensors and actuating the valves. Data analysis was conducted using MATLAB (MATLAB 2022a, MathWorks, USA). The monitoring and control stations served as external elements to the robot prototypes.

3. Results
3.1 Earthworm-inspired robot

The innovative design of the earthworm-inspired robot is based on a single linear actuator. Here, locomotion was achieved by the elongation and contraction of the actuator, when inflated and deflated. It was engineered together with the effect of the state-dependent friction between the convex shape of its legs and the flat ground. The actuator was placed horizontally, supported by two structures connected to the back and front legs. These structures were guided by two axes, ensuring their parallel alignment, while the structure of the front leg could slide along it, Fig. 4A.

The end of the leg that comes into contact with the flat ground is convex, and each half is made of two different materials, an elastomer and a rigid material. Friction is state-dependent because when the convex shape in contact with the ground is the elastomer, we have relatively higher friction than when it is the rigid material. The section of the convex shape in contact with the ground changes with the deflection of the legs, which, in turn, changes with the elongation and contraction of the actuator. When the actuator elongates, the back leg deflects so that the higher friction section is in contact with the ground, while the front leg deflects to have the lower friction section in contact with the ground, Fig. 4B. This results in a higher friction force on the back leg than on the front leg, $F_{b/f} > F_{f/f}$. Since the elongation/contraction force $F_{e/c}$ is the same at both the back and front legs, the resultant force $F_r$ is the difference between both friction forces. Consequently, this causes the front leg to slip on the ground while the back leg remains fixed. Similar reasoning can be applied during the contraction of the actuator. In this scenario, the higher friction section is on the front leg, leading in a higher friction force on the front leg, $F_{b/f} < F_{f/f}$. As a result, the back leg slips on the ground while the front leg remains fixed. The resultant force in both elongation and contraction points forward, promoting robot locomotion (Movie S3, Supplementary Materials).

Locomotion speed depends on a range of factors, including the dimensions of the actuator, relative friction, and the control process, particularly the actuation frequency. An on-off pneumatic control cycle was implemented for inflating/deflating the actuator. The robot's locomotion was assessed by varying the actuation frequency, reaching its maximum speed at 0.8 Hz (~16 mm/s) with a constant pressure of 40 kPa, Fig. 4C. Control frequencies higher than 1.2 Hz do not allow sufficient time for the actuator to fully depressurize, resulting in an incomplete inflating/deflating cycle. This is visible in the reduced amplitude of movement, characterized by smaller and faster steps, ultimately resulting in the inability of the robot to move. The depressurization time is influenced by the design and material of the actuator, particularly the force exerted by the actuator to expel the air inside the chamber during the deflation phase of the cycle.

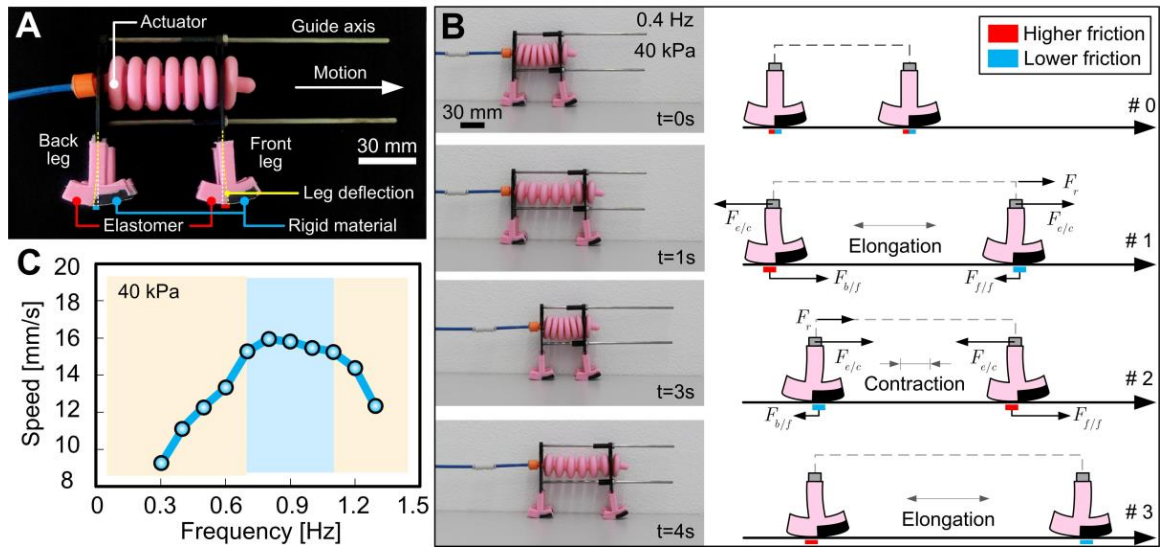

**Fig. 4**. Earthworm-inspired soft robot. (A) Robot design composed of a linear pneumatic actuator and flexural legs with a convex shaped end made of two different materials. (B) Locomotion is generated by the elongation/contraction of the actuator combined with the state-dependent friction between the convex ends of its legs and the ground. (C) The effect of the on-off control frequency on locomotion speed. The speed was assessed by analyzing recorded videos varying the on-off control frequency.

### 3.2 Four-legged robot

Robots are increasingly used for exploration on land, underwater, and even in space. While navigating, robot wheels face limitations in overcoming all obstacles, and legs made of rigid materials are challenging to control in unstructured ground conditions. In contrast, soft-legged robots can navigate through uncertain terrain, absorbing unexpected collisions, and ensuring safety when sharing workspaces with people and equipment [69, 70].

Here, we combined four bending #2 type actuators, each acting as a statically stable leg, to create a four-legged robot capable of locomotion across different terrains. The locomotion cycle is run by an on-off closed-loop controller that inflates/deflates the robot's legs. The robot's legs are actuated in an antisymmetric pattern, with the front left leg (FL) and the back right leg (BR) simultaneously pressurized as one set, while the other set is composed of the front right leg (FR) and the back left leg (BL), Fig. 5A. A complete cycle of the actuation pattern takes 900 ms. FEA numerical results established the balance between the input pressures and the resulting bendability of the actuators. The wall thickness of 4 mm was empirically defined to ensure that four actuators can hold their position under static conditions while supporting added loads.

We tested and evaluated the robot's locomotion capabilities in two different environments, an unstructured garden area and a flat surface. The tests in the garden demonstrated the feasibility of locomotion even on sloped ground. However, its performance is constrained by the relatively small amplitude of the legs' movement, as well as their relatively small size compared to obstacles on the ground, Fig. 5B and Movie S4, Supplementary Materials. We evaluated the robot's speed with varying control pressures and added loads to assess the feasibility of locomotion on a flat surface. Higher input pressures lead to an increase in the size of each step, which, in turn, resulted in faster robot speeds, Fig. 5C and Movie S4, Supplementary Materials. Based on data from FEA, we defined a maximum pressure of 50 kPa to balance robot speed with the mechanical operability of the actuators. We also tested the robot speed by fixing the input pressure at 50 kPa and adding different masses on top of the robot. As the load masses increased, the size of the robot's steps was reduced, resulting in a decrease in speed, Fig. 5D and Movie S4, Supplementary Materials. The walking speed was evaluated by analyzing the recorded videos, with variations in the input pressure of the actuators (3 trials were conducted for each pressure). The same process was applied to estimate the robot walking speed as a function of the load masses it carries at a constant pressure of 50 kPa.

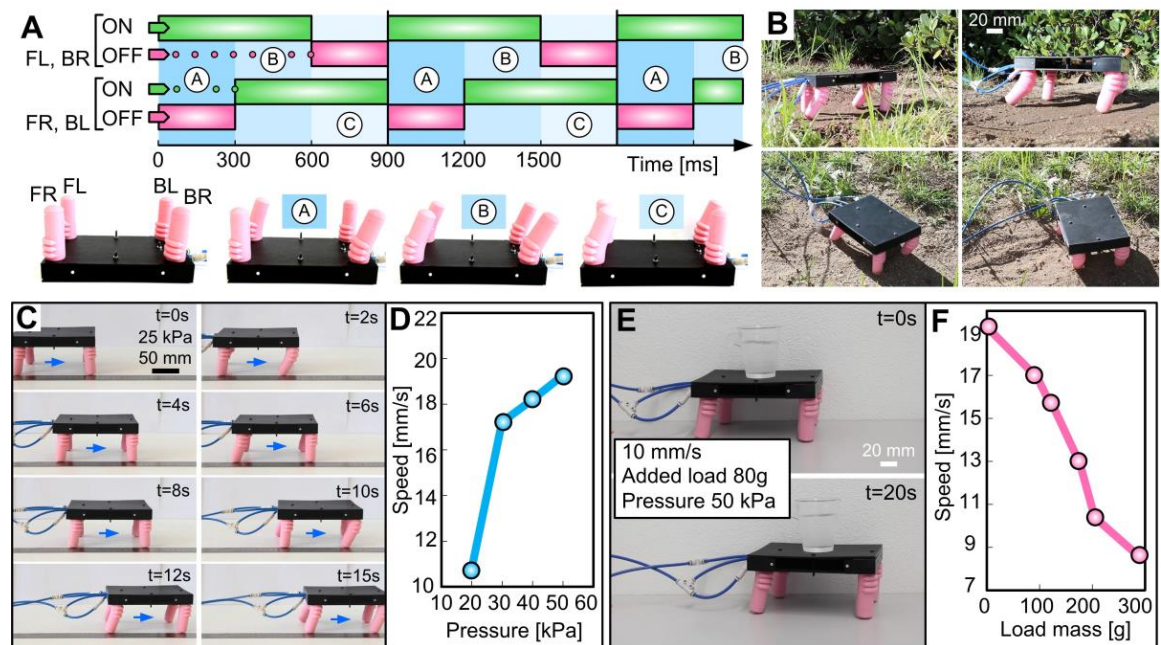

**Fig. 5**. Conceptual design of a four-legged robot inspired by quadruped animals. (A) Antisymmetric locomotion actuation pattern and related on-off signal diagram for each leg. The FL and BR legs are paired and operate in coordination, as do the FR and BL legs. (B) The robot walking on an unstructured garden flat surface. (C) Robot walking at various time intervals under an inner pressure of 25 kPa. (D) Robot walking speed as function of the inner pressure. (E) Robot walking at a speed of 10 mm/s with an added load of 80 grams. (F) Robot walking speed as a function of the load masses it carried at a constant pressure of 50 kPa.

### 3.3 Gripper

We combined three bending #1 actuators to construct a robot gripper designed for grasping objects of varying weight, size, geometry, and surface texture from the Yale-CMU-Berkeley (YCB) object and model set [71], Fig. 6A. The gripper, attached to a robot flange, demonstrated its ability to grasp, lift, hold (for 10 s), and release various objects. It adjusted the contact area with them according to the actuation pressure, size and geometry of the object, Fig. 6B and Movie S5, Supplementary Materials.

Results indicated that the actuation pressure needed for grasping-lifting-holding-releasing the objects was relatively high based on the pressure data from the FEA. At lower pressures, noticeable slippage between the actuators and the objects was observed, Fig. 6C. We addressed this challenge by adding a strip of nano/gecko tape to the contact surface of each actuator. As a result, the same objects could be grasped-lifted-held-released at significantly lower pressures (~65% less), Fig. 6C. Due to the intricate geometry of the Rubik's cube and the relatively high mass of the baseball, both objects were exclusively grasped-lifted-held-released using nano tape. The nano tape used was a 20×5 mm strip with 29,000 fastening elements per cm2 (Gecko® Tape, blinder, Germany).

We also varied the mass of the spherical-shaped object by adding different quantities of steel filings inside to achieve the desired mass values. The sphere diameter was optimized to maximize the contact area with the soft actuators. The sphere's surface texture resulted from its FFF-based printing in rigid material. As seen in previous results, higher pressures were required to grasp-lift-hold-release the sphere without using nano tape, limiting the maximum mass to 200 g, Fig. 6D. On the other hand, using nano tape, the gripper successfully grasped-lifted-held-released a sphere weighing 267 g, approximately five times heavier than the gripper's own weight of 54 g. Pressures higher than 40 kPa did not affect the gripper's performance. The gripper's actuation pressures were measured using an absolute pressure sensor (MPX4250AP, NXP Semiconductors, Netherlands), and the masses of the objects were determined using a weighing scale (ES-3000A, TechMaster, USA). The reported pressure and mass values represent the averages of 3 trials conducted for each object.

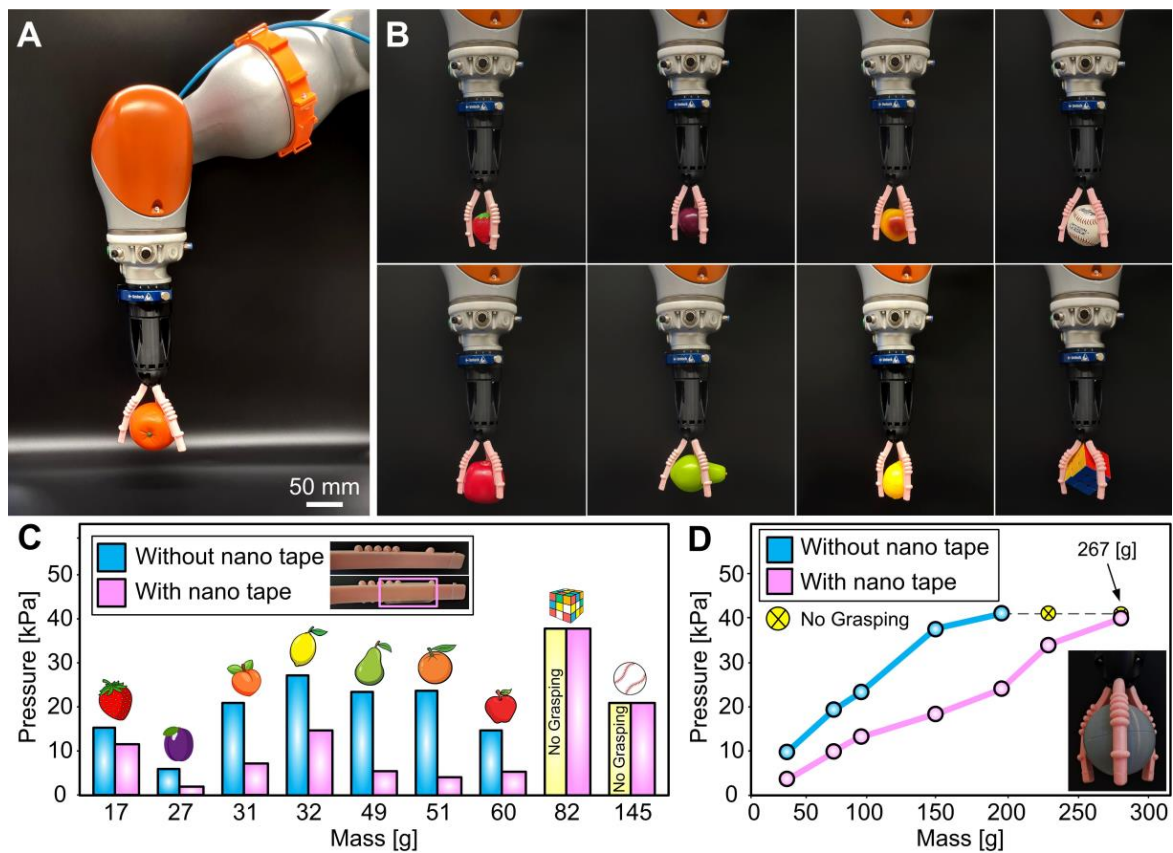

**Fig. 6**. Soft gripper grasp-lift-hold-release testing and evaluation. (A) The soft gripper, attached to a robot, can grasp-lift-hold-release an orange from the YCB object and model set. (B) The gripper grasping-lifting-holding-releasing multiple objects with varying weight, size, geometry, and surface texture. (C) Minimum actuation pressure required to grasp-lift-hold-release YCB objects featuring different masses, with results shown for both using and not using the nano tape. (D) Actuation pressure required to grasp-lift-hold-

release a spherical-shaped object of varied mass, with results presented for both using and not using nano tape.

## 4. Conclusion and discussion

We presented an integrated process for the design and fabrication of soft robot actuators in a single casting in this work. The FEA effectively assisted in ensuring the mechanical operability and functionality of the actuators, allowing us to anticipate the effects of different input pressures on their elongation and bending. Moreover, FEA-assisted design eliminated the lengthy and costly trial-and-error design-fabrication processes, which often leads to the fabrication of multiple prototypes.

The fabrication process in a single casting step relies on an effective methodology to fully remove the sacrificial water-soluble cores. The heated water circuit accelerates the dissolution of the material constituting the core and ensures its complete removal from the actuator's walls. The design of the core, water temperature, and water flow through the hollow core are key elements for the successful fabrication of the actuators. This process facilitated the fabrication of actuators with non-uniform cross sections geometry under minimal supervision and at an affordable cost. We demonstrated the feasibility of this approach using a desktop FFF printer, a vacuum chamber, and a heated water circuit, with a total equipment cost of less than 1,000 dollars. The materials cost for each actuator of any type is approximately 1 dollar, including the molds, core, and silicone. The control elements for each actuator cost around 15 dollars, including 2 valves, a pressure sensor, and tubes. The control board can simultaneously control multiple actuators and costs about 35 dollars. The estimated cost of the earthworm-inspired robot is approximately 55 dollars, while the cost of the four-legged robot is around 100 dollars. The gripper uses a single control element for the three fingers with a total cost of about 60 dollars. After the silicone is cured, and considering that the supporting elements are available, each robot prototype can be finished in less than 3 hours. In conclusion, the proposed fabrication process boasts high repeatability, consistently producing quality functional and reliable actuators. Moreover, it relies on readily available equipment and materials, comes at a reduced cost, requires minimal skilled labor and processing steps/handling.

The design and fabrication of three distinct pneu-net inspired actuators have led to the development of three soft robot prototypes, two legged robots and a robotic gripper. The homogeneous or heterogeneous combination of these actuators has the potential to accelerate the design and fabrication of innovative soft robots, making them more accessible to people at a lower cost.

The integrated design and fabrication method proposed here has demonstrated high effectiveness. However, it faces challenges that could potentially hinder the fabrication of functional actuators. The reliability of FEA results is highly dependent on the hyperelastic material model, involving iterative procedures that compare numerical predictions with physical measurements to calibrate the system. These procedures can be time-consuming as the physical measurements are obtained from physical models that have to be fabricated previously. In the future, the hyperelastic material models will likely and automatically adapt, not only to the material's properties but also to the geometry of the actuators. The utilization of a desktop FFF printer for fabricating sacrificial cores can introduce geometrical inaccuracies in the actuators fabricated, resulting in non-uniform wall thickness and surface defects in the wall chambers. These inaccuracies and defects stem from an improper definition of the printing parameters, which are highly dependent on the storage conditions of the water-soluble filament and the printing environment. In the future, the printing of sacrificial cores could be achieved using alternative water-soluble materials that are less dependent on storage and printing conditions. Good quality cores could be printed alongside the molds using desktop multi-nozzle FFF machines to maintain the low-cost nature of the system.

## Acknowledgments

**General**: We thank P. Matos for the valuable discussions and suggestions on the fabrication of soft actuators. We thank S. Alves for the valuable suggestions on the finite element analysis of hyperelastic material.

**Author contributions:** A.S., D.F. and M.B. contributed to the design conceptualization, fabrication, and conducted the experimental tests. D.N. contributed to the numerical simulation. D.F. and M.B. contributed to the analysis and result interpretation, and P.N. contributed to funding acquisition, review and writing.

**Funding:** This work was supported by Portuguese national funds through FCT - Fundação para a Ciência e a Tecnologia, [grant numbers UIDB/00285/2020, LA/P/0112/2020 and 2022.13512.BD].

**Competing interests:** The authors declare that there is no conflict of interest regarding the publication of this article.

**Data Availability:** The data that support the findings of this study are available from the corresponding author upon reasonable request.

## Supplementary Materials

Figure S1 to S2
Movies S1 to S5

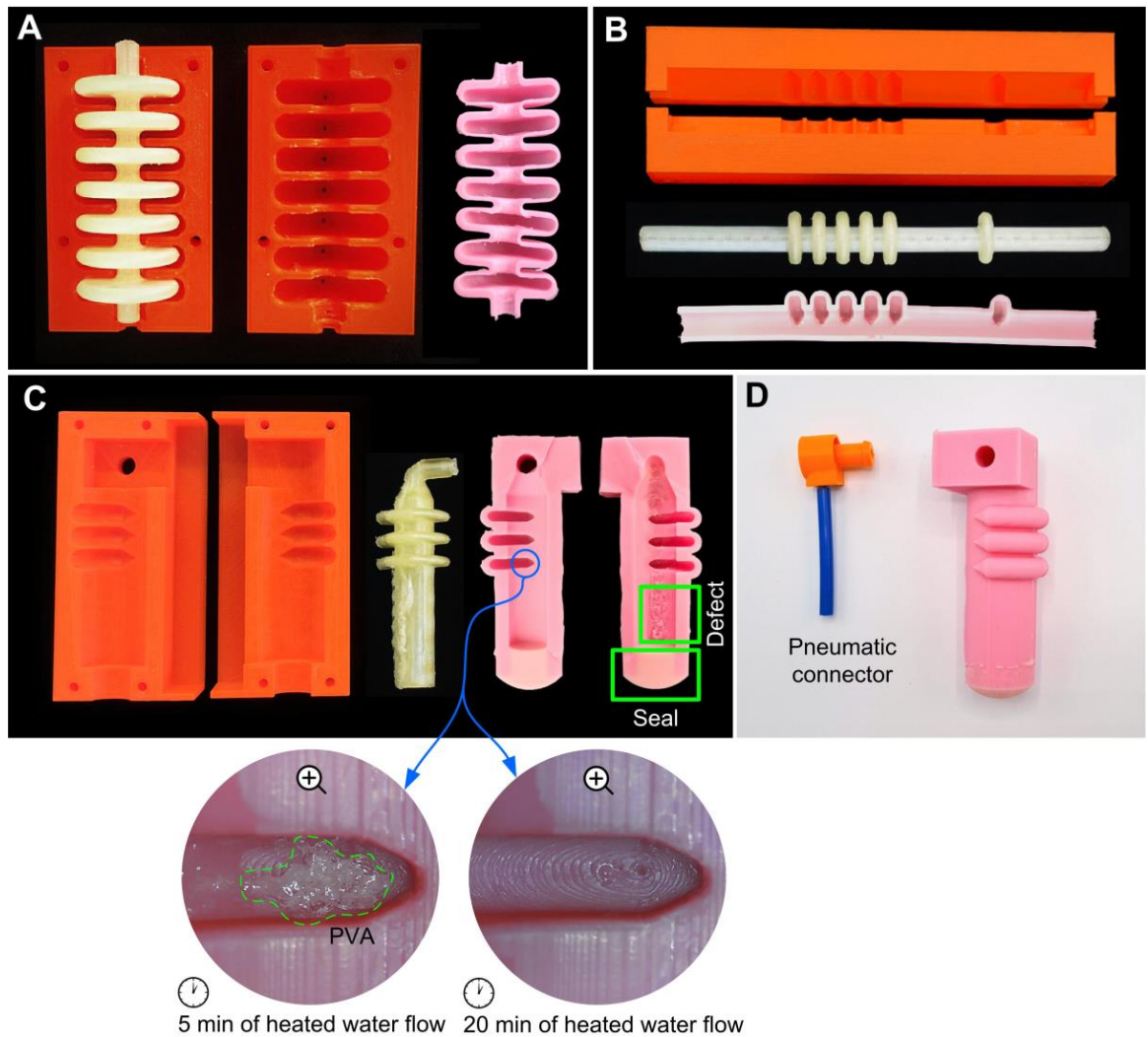

**Fig. S1**. Molding elements and fabricated actuators. (A) Molds and core of the linear actuator. The section view of the actuator reveals a non-uniform wall thickness due to geometrical inaccuracies in the core. (B) Molds and core of the bending #1 actuator, along with the section view of the actuator. (C) Molds and core of the bending #2 actuator. The actuator section view displays surface defects in the chamber walls. The zoomed-in images show that in intricate chamber geometry some PVA residue is visible when the water is circulated for only 5 minutes. However, when the water is circulated for 20 minutes, the void is completely free of unwanted residue. (D) Finished bending #2 actuator with the pneumatic connector. Both non-uniform wall thickness and surface defects do not compromise the actuators functionality.

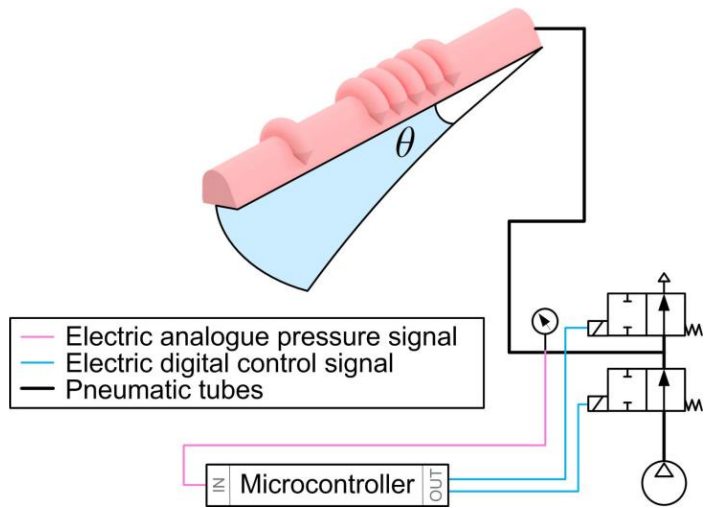

**Fig. S2**. Actuator monitoring and control elements. Two solenoid 2/2 valves control each actuator, with one functioning as an escape valve. Each actuating system is monitored by a single pressure sensor. The control lines are in blue and the monitoring lines in pink.

**Movie S1**: FABRICATION

The fabrication of the pneu-net inspired soft actuators follows a sequential process. Molds and sacrificial cores are fabricated using desktop 3D printers (FFF technology). The molds are made of polylactic acid (PLA) and the hollow cores are made of water-soluble polyvinyl alcohol (PVA). The molds and core are assembled, following the single step casting process where the elastomer in liquid-viscous state is poured into the molds. After ~2 minutes in the vacuum chamber to remove air bubbles and 24 hours of curing at room temperature, the unfinished actuator (with the core inside) is extracted from the mold. The actuator is placed at the heated water circuit to remove the sacrificial water-soluble core. The pump forces the circulation of filtered water through the hollow core at a temperature of ~65 ºC.

**Movie S2**: FEA-AIDED DESIGN

Single chamber actuators with non-uniform cross section geometries are designed in CAD. The finite element analysis (FEA), using Inventor Nastran, aided in the optimization of the actuator design and mechanical operability. The relationship between the input pressure and the deformed configuration of the actuator is analyzed using the constitutive model Neo-Hookean hyperelastic. After optimizing the actuator design, its CAD model is used to support the design of the holed core and molds.

**Movie S3**: EARTHWORM-INSPIRED ROBOT

The earthworm-inspired robot design is based on a single linear actuator. Locomotion is achieved by the elongation and contraction of the actuator, together with the effect of the state-dependent friction between the convex shape of its legs and the flat ground. During elongation, the front leg slips on the ground while the back leg remains fixed, while during contraction the back leg slips on the ground while the front leg remains fixed. The on-off pneumatic control cycle resulted in a robot maximum speed at 0.8 Hz (~16 mm/s) with an input pressure of 40 kPa. For control frequencies higher than 1.2 Hz there is no locomotion as there is not sufficient time for the actuator to fully depressurize.

**Movie S4**: FOUR-LEGGED ROBOT LOCOMOTION

The four-legged robot design is inspired by quadruped animals. The robot is composed of four bending #2 actuators acting as statically stable legs. Locomotion is achieved by implementing an antisymmetric locomotion actuation pattern. The front left (FL) and back right (BR) legs are paired and operate in coordination. The same principle applies to the front right (FR) and back left (BL) legs. The walking speed is a function of the inner pressure within the actuator's chamber and the load masses that the robot is carrying. Locomotion in the unstructured garden is constrained by the relatively small amplitude of the legs and their relatively small size compared to obstacles on the ground.

**Movie S5**: SOFT GRIPPER

The soft gripper is composed of three bending #1 actuators. It was attached to the flange of a robot manipulator (iiwa, KUKA, Germany), to grasp, lift, hold (for 10 s), and release objects of varying weight, size, geometry, and surface texture from the Yale-CMU-Berkeley (YCB) object and model set. Results show that using nano/gecko tape the objects could be grasped-lifted-held-released at significantly lower pressures (~65% less). The Rubik's cube and the baseball were exclusively grasped-lifted-held-released using nano/gecko tape. Using nano/gecko tape, the gripper successfully grasped-lifted-held-released objects five times heavier than the gripper's own weight. Pressures higher than 40 kPa did not affect the gripper's performance.

**References**

1. J. K. A. Langowski, P. Sharma, A. L. Shoushtari, In the soft grip of nature. *Sci. Robot.* **5** (2020).

2. Z. Hu, Y. Zhang, H. Jiang, J. Lv, Bioinspired helical-artificial fibrous muscle structured tubular soft actuators. *Sci. Adv.* **9** (2023).

3. W. Kim, J. Byun, J.-K. Kim, W.-Y. Choi, K. Jakobsen, J. Jakobsen, D.-Y. Lee, K.-J. Cho, Bioinspired dual-morphing stretchable origami. *Sci. Robot.* **4**, 1–11 (2019).

4. S. Wu, Y. Hong, Y. Zhao, J. Yin, Y. Zhu, Caterpillar-inspired soft crawling robot with distributed programmable thermal actuation. *Sci. Adv.* **9** (2023).

5. H. Zhou, S. Cao, S. Zhang, F. Li, N. Ma, Design of a Fuel Explosion-Based Chameleon-Like Soft Robot Aided by the Comprehensive Dynamic Model. *Cyborg Bionic Syst.* **4**, 1–14 (2023).

6. C. Laschi, B. Mazzolai, Lessons from Animals and Plants: The Symbiosis of Morphological Computation and Soft Robotics. *IEEE Robot. Autom. Mag.* **23**, 107–114 (2016).

7. T. J. Wallin, J. Pikul, R. F. Shepherd, 3D printing of soft robotic systems. *Nat. Rev. Mater.* **3**, 84–100 (2018).

8. C. Laschi, B. Mazzolai, M. Cianchetti, Soft robotics: Technologies and systems pushing the boundaries of robot abilities. *Sci. Robot.* **1**, 1–11 (2016).

9. W. Dou, G. Zhong, J. Cao, Z. Shi, B. Peng, L. Jiang, Soft Robotic Manipulators: Designs, Actuation, Stiffness Tuning, and Sensing. *Adv. Mater. Technol.* **6**, 1–24 (2021).

10. M. Wehner, R. L. Truby, D. J. Fitzgerald, B. Mosadegh, G. M. Whitesides, J. A. Lewis, R. J. Wood, An integrated design and fabrication strategy for entirely soft, autonomous robots. *Nature* **536**, 451–455 (2016).

11. D. Rus, M. T. Tolley, Design, fabrication and control of soft robots. *Nature* **521**, 467–475 (2015).

12. C. Huang, Z. Lai, X. Wu, T. Xu, Multimodal Locomotion and Cargo Transportation of Magnetically Actuated Quadruped Soft Microrobots. *Cyborg Bionic Syst.* **2022**, 0004 (2022).

13. S. M. Youssef, M. Soliman, M. A. Saleh, M. A. Mousa, M. Elsamanty, A. G. Radwan, Underwater Soft Robotics: A Review of Bioinspiration in Design, Actuation, Modeling, and Control. *Micromachines* **13**, 110 (2022).

14. M. Runciman, A. Darzi, G. P. Mylonas, Soft Robotics in Minimally Invasive Surgery. *Soft Robot.* **6**, 423–443 (2019).

15. E. W. Hawkes, L. H. Blumenschein, J. D. Greer, A. M. Okamura, A soft robot that navigates its environment through growth. *Sci. Robot.* **2**, 1–8 (2017).

16. S. Zhong, Z. Xin, Y. Hou, Y. Li, H.-W. Huang, T. Sun, Q. Shi, H. Wang, Double-Modal Locomotion of a Hydrogel Ultra-Soft Magnetic Miniature Robot with Switchable Forms. *Cyborg Bionic Syst.* **5** (2024).

17. P. Rothemund, A. Ainla, L. Belding, D. J. Preston, S. Kurihara, Z. Suo, G. M. Whitesides, A soft, bistable valve for autonomous control of soft actuators. *Sci. Robot.* **3**, 1–11 (2018).

18. N. W. Bartlett, K. P. Becker, R. J. Wood, A fluidic demultiplexer for controlling large arrays of soft actuators. *Soft Matter* **16**, 5871–5877 (2020).

19. M. S. Xavier, A. J. Fleming, Y. K. Yong, Finite Element Modeling of Soft Fluidic Actuators: Overview and Recent Developments. *Adv. Intell. Syst.* **3**, 2000187 (2021).

20. V. Subramaniam, S. Jain, J. Agarwal, P. Valdivia y Alvarado, Design and characterization of a hybrid soft gripper with active palm pose control. *Int. J. Rob. Res.* **39**, 1668–1685 (2020).

21. L. Marechal, P. Balland, L. Lindenroth, F. Petrou, C. Kontovounisios, F. Bello, Toward a Common Framework and Database of Materials for Soft Robotics. *Soft Robot.* **8**, 284–297 (2021).

22. S. Alves, M. Babcinschi, A. Silva, D. Neto, D. Fonseca, P. Neto, Integrated Design Fabrication and Control of a Bioinspired Multimaterial Soft Robotic Hand. *Cyborg Bionic Syst.* **4**, 1–10 (2023).

23. A. K. Mishra, F. Tramacere, B. Mazzolai, "From plant root's sloughing and radial expansion mechanisms to a soft probe for soil exploration" in *2018 IEEE International Conference on Soft Robotics (RoboSoft)* (IEEE, 2018; https://ieeexplore.ieee.org/document/8404899/), pp. 71–76.

24. F. Schmitt, O. Piccin, L. Barbé, B. Bayle, Soft Robots Manufacturing: A Review. *Front. Robot. AI* **5** (2018).

25. H. Li, J. Yao, P. Zhou, X. Chen, Y. Xu, Y. Zhao, High-force soft pneumatic actuators based on novel casting method for robotic applications. *Sensors Actuators A Phys.* **306**,

111957 (2020).

26. B. Mazzolai, A. Mondini, F. Tramacere, G. Riccomi, A. Sadeghi, G. Giordano, E. Del Dottore, M. Scaccia, M. Zampato, S. Carminati, Octopus-Inspired Soft Arm with Suction Cups for Enhanced Grasping Tasks in Confined Environments. *Adv. Intell. Syst.* **1**, 1900041 (2019).

27. T. J. Jones, E. Jambon-Puillet, J. Marthelot, P.-T. Brun, Bubble casting soft robotics. *Nature* **599**, 229–233 (2021).

28. T. Dahlberg, T. Stangner, H. Zhang, K. Wiklund, P. Lundberg, L. Edman, M. Andersson, 3D printed water-soluble scaffolds for rapid production of PDMS micro-fluidic flow chambers. *Sci. Rep.* **8**, 3372 (2018).

29. W. H. Goh, M. Hashimoto, Fabrication of 3D Microfluidic Channels and In-Channel Features Using 3D Printed, Water-Soluble Sacrificial Mold. *Macromol. Mater. Eng.* **303**, 1–9 (2018).

30. M. Alsharari, B. Chen, W. Shu, Sacrificial 3D Printing of Highly Porous, Soft Pressure Sensors. *Adv. Electron. Mater.* **8**, 1–12 (2022).

31. V. Saggiomo, A. H. Velders, Simple 3D Printed Scaffold-Removal Method for the Fabrication of Intricate Microfluidic Devices. *Adv. Sci.* **2**, 1–5 (2015).

32. H.-Y. Chen, A. T. Conn, A Stretchable Inductor With Integrated Strain Sensing and Wireless Signal Transfer. *IEEE Sens. J.* **20**, 7384–7391 (2020).

33. N. W. Bartlett, M. T. Tolley, J. T. B. Overvelde, J. C. Weaver, B. Mosadegh, K. Bertoldi, G. M. Whitesides, R. J. Wood, A 3D-printed, functionally graded soft robot powered by combustion. *Science (80-. ).* **349**, 161–165 (2015).

34. J. D. Hubbard, R. Acevedo, K. M. Edwards, A. T. Alsharhan, Z. Wen, J. Landry, K. Wang, S. Schaffer, R. D. Sochol, Fully 3D-printed soft robots with integrated fluidic circuitry. *Sci. Adv.* **7** (2021).

35. R. L. Truby, J. A. Lewis, Printing soft matter in three dimensions. *Nature* **540**, 371–378 (2016).

36. A. Zolfagharian, A. Z. Kouzani, S. Y. Khoo, A. A. A. Moghadam, I. Gibson, A. Kaynak, Evolution of 3D printed soft actuators. *Sensors Actuators A Phys.* **250**, 258–272 (2016).

37. S. R. Dabbagh, M. R. Sarabi, M. T. Birtek, S. Seyfi, M. Sitti, S. Tasoglu, 3D-printed microrobots from design to translation. *Nat. Commun.* **13**, 5875 (2022).

38. X. Liang, Z. Chen, Y. Deng, D. Liu, X. Liu, Q. Huang, T. Arai, Field-Controlled Microrobots Fabricated by Photopolymerization. *Cyborg Bionic Syst.* **4**, 1–14 (2023).

39. Y. Zhang, C. J. Ng, Z. Chen, W. Zhang, S. Panjwani, K. Kowsari, H. Y. Yang, Q. Ge, Miniature Pneumatic Actuators for Soft Robots by High-Resolution Multimaterial 3D Printing. *Adv. Mater. Technol.* **4**, 1–9 (2019).

40. R. D. Sochol, E. Sweet, C. C. Glick, S. Venkatesh, A. Avetisyan, K. F. Ekman, A. Raulinaitis, A. Tsai, A. Wienkers, K. Korner, K. Hanson, A. Long, B. J. Hightower, G. Slatton, D. C. Burnett, T. L. Massey, K. Iwai, L. P. Lee, K. S. J. Pister, L. Lin, 3D printed microfluidic circuitry via multijet-based additive manufacturing. *Lab Chip* **16**, 668–678 (2016).

41. H. Jia, J. Flommersfeld, M. Heymann, S. K. Vogel, H. G. Franquelim, D. B. Brückner, H. Eto, C. P. Broedersz, P. Schwille, 3D printed protein-based robotic structures actuated by molecular motor assemblies. *Nat. Mater.* **21**, 703–709 (2022).

42. C. De Pascali, G. A. Naselli, S. Palagi, R. B. N. Scharff, B. Mazzolai, 3D-printed biomimetic artificial muscles using soft actuators that contract and elongate. *Sci. Robot.* **7** (2022).

43. Y. Zhai, A. De Boer, J. Yan, B. Shih, M. Faber, J. Speros, R. Gupta, M. T. Tolley, Desktop fabrication of monolithic soft robotic devices with embedded fluidic control circuits. *Sci. Robot.* **8**, 1–14 (2023).

44. Y. Sun, Y. Liu, F. Pancheri, T. C. Lueth, LARG: A Lightweight Robotic Gripper With 3-D Topology Optimized Adaptive Fingers. *IEEE/ASME Trans. Mechatronics* **27**, 2026–2034 (2022).

45. A. K. Mishra, T. J. Wallin, W. Pan, A. Xu, K. Wang, E. P. Giannelis, B. Mazzolai, R. F. Shepherd, Autonomic perspiration in 3D-printed hydrogel actuators. *Sci. Robot.* **5**, 1–9 (2020).

46. A. Haake, R. Tutika, G. M. Schloer, M. D. Bartlett, E. J. Markvicka, On-Demand Programming of Liquid Metal-Composite Microstructures through Direct Ink Write 3D Printing. *Adv. Mater.* **34** (2022).

47. J. Kwon, C. DelRe, P. Kang, A. Hall, D. Arnold, I. Jayapurna, L. Ma, M. Michalek, R. O. Ritchie, T. Xu, Conductive Ink with Circular Life Cycle for Printed Electronics. *Adv. Mater.* **34**, 1–7 (2022).

48. J. W. Boley, E. L. White, G. T. C. Chiu, R. K. Kramer, Direct Writing of Gallium-Indium Alloy for Stretchable Electronics. *Adv. Funct. Mater.* **24**, 3501–3507 (2014).

49. A. D. Valentine, T. A. Busbee, J. W. Boley, J. R. Raney, A. Chortos, A. Kotikian, J. D. Berrigan, M. F. Durstock, J. A. Lewis, Hybrid 3D Printing of Soft Electronics. *Adv. Mater.* **29**, 1–8 (2017).

50. M. A. Darabi, A. Khosrozadeh, R. Mbeleck, Y. Liu, Q. Chang, J. Jiang, J. Cai, Q. Wang, G. Luo, M. Xing, Skin-Inspired Multifunctional Autonomic-Intrinsic Conductive Self-Healing Hydrogels with Pressure Sensitivity, Stretchability, and 3D Printability. *Adv. Mater.* **29**, 1–8 (2017).

51. R. L. Truby, M. Wehner, A. K. Grosskopf, D. M. Vogt, S. G. M. Uzel, R. J. Wood, J. A. Lewis, Soft Somatosensitive Actuators via Embedded 3D Printing. *Adv. Mater.* **30**, 1–8 (2018).

52. Y. Kim, H. Yuk, R. Zhao, S. A. Chester, X. Zhao, Printing ferromagnetic domains for untethered fast-transforming soft materials. *Nature* **558**, 274–279 (2018).

53. A. Kotikian, R. L. Truby, J. W. Boley, T. J. White, J. A. Lewis, 3D Printing of Liquid Crystal Elastomeric Actuators with Spatially Programed Nematic Order. *Adv. Mater.* **30**, 1–6 (2018).

54. B. Hayes, T. Hainsworth, R. MacCurdy, Liquid–solid co-printing of multi-material 3D fluidic devices via material jetting. *Addit. Manuf.* **55**, 102785 (2022).

55. R. MacCurdy, R. Katzschmann, Youbin Kim, D. Rus, "Printable hydraulics: A method for fabricating robots by 3D co-printing solids and liquids" in *2016 IEEE International Conference on Robotics and Automation (ICRA)* (IEEE, 2016; http://ieeexplore.ieee.org/document/7487576/)vols. 2016-June, pp. 3878–3885.

56. M. A. Skylar-Scott, J. Mueller, C. W. Visser, J. A. Lewis, Voxelated soft matter via multimaterial multinozzle 3D printing. *Nature* **575**, 330–335 (2019).

57. T. Hainsworth, L. Smith, S. Alexander, R. MacCurdy, A Fabrication Free, 3D Printed, Multi-Material, Self-Sensing Soft Actuator. *IEEE Robot. Autom. Lett.* **5**, 4118–4125 (2020).

58. Z. X. Khoo, J. E. M. Teoh, Y. Liu, C. K. Chua, S. Yang, J. An, K. F. Leong, W. Y. Yeong, 3D printing of smart materials: A review on recent progresses in 4D printing. *Virtual Phys. Prototyp.* **10**, 103–122 (2015).

59. C. Cvetkovic, R. Raman, V. Chan, B. J. Williams, M. Tolish, P. Bajaj, M. S. Sakar, H. H. Asada, M. T. A. Saif, R. Bashir, Three-dimensionally printed biological machines powered by skeletal muscle. *Proc. Natl. Acad. Sci.* **111**, 10125–10130 (2014).

60. D. L. Taylor, M. in het Panhuis, Self-Healing Hydrogels. *Adv. Mater.* **28**, 9060–9093 (2016).

61. D. K. Patel, A. H. Sakhaei, M. Layani, B. Zhang, Q. Ge, S. Magdassi, Highly Stretchable and UV Curable Elastomers for Digital Light Processing Based 3D Printing. *Adv. Mater.* **29**, 1–7 (2017).

62. M. Zadan, D. K. Patel, A. P. Sabelhaus, J. Liao, A. Wertz, L. Yao, C. Majidi, Liquid Crystal Elastomer with Integrated Soft Thermoelectrics for Shape Memory Actuation and Energy Harvesting. *Adv. Mater.* **34** (2022).

63. X. Kuang, K. Chen, C. K. Dunn, J. Wu, V. C. F. Li, H. J. Qi, 3D Printing of Highly Stretchable, Shape-Memory, and Self-Healing Elastomer toward Novel 4D Printing. *ACS Appl. Mater. Interfaces* **10**, 7381–7388 (2018).

64. H. K. Yap, H. Y. Ng, C.-H. Yeow, High-Force Soft Printable Pneumatics for Soft Robotic Applications. *Soft Robot.* **3**, 144–158 (2016).

65. B. A. W. Keong, R. Y. C. Hua, A Novel Fold-Based Design Approach toward Printable Soft Robotics Using Flexible 3D Printing Materials. *Adv. Mater. Technol.* **3**, 1–12 (2018).

66. S. E. Bakarich, M. In Het Panhuis, S. Beirne, G. G. Wallace, G. M. Spinks, Extrusion printing of ionic–covalent entanglement hydrogels with high toughness. *J. Mater. Chem. B* **1**, 4939 (2013).

67. X. Cao, S. Xuan, Y. Gao, C. Lou, H. Deng, X. Gong, 3D Printing Ultraflexible Magnetic Actuators via Screw Extrusion Method. *Adv. Sci.* **9**, 1–12 (2022).

68. B. Mosadegh, P. Polygerinos, C. Keplinger, S. Wennstedt, R. F. Shepherd, U. Gupta, J. Shim, K. Bertoldi, C. J. Walsh, G. M. Whitesides, Pneumatic Networks for Soft Robotics that Actuate Rapidly. *Adv. Funct. Mater.* **24**, 2163–2170 (2014).

69. D. Drotman, S. Jadhav, D. Sharp, C. Chan, M. T. Tolley, Electronics-free pneumatic circuits for controlling soft-legged robots. *Sci. Robot.* **6** (2021).

70. S. T. Muralidharan, R. Zhu, Q. Ji, L. Feng, X. V. Wang, L. Wang, "A soft quadruped robot enabled by continuum actuators" in *2021 IEEE 17th International Conference on Automation Science and Engineering (CASE)* (IEEE, 2021; https://ieeexplore.ieee.org/document/9551496/)vols. 2021-Augus, pp. 834–840.

71. B. Calli, A. Walsman, A. Singh, S. Srinivasa, P. Abbeel, A. M. Dollar, Benchmarking in Manipulation Research: Using the Yale-CMU-Berkeley Object and Model Set. *IEEE Robot. Autom. Mag.* **22**, 36–52 (2015).